\documentclass[10pt,journal,twoside,web]{ieeecolor}
\pdfoutput=1
\usepackage{generic}
\usepackage{cite}
\usepackage{amsmath,amssymb,amsfonts}
\usepackage{algorithmic}
\usepackage{graphicx}
\usepackage{textcomp}
\usepackage{multirow}
\usepackage{soul}
\usepackage{comment}
\usepackage{hyperref}
\usepackage[outdir=./]{epstopdf}
\usepackage[T1]{fontenc}
\usepackage[utf8]{inputenc}
\usepackage{hyperref}
\usepackage{booktabs}
\usepackage{multicol}
\usepackage{multirow}

\def\BibTeX{{\rm B\kern-.05em{\sc i\kern-.025em b}\kern-.08em
    T\kern-.1667em\lower.7ex\hbox{E}\kern-.125emX}}
\def\BibTeX{{\rm B\kern-.05em{\sc i\kern-.025em b}\kern-.08em
    T\kern-.1667em\lower.7ex\hbox{E}\kern-.125emX}}

\begin{document}

\title{Automated Sleep Staging via Parallel Frequency-Cut Attention}
\author{Zheng Chen$^{\dagger}$,  Ziwei Yang, Lingwei Zhu,
Wei Chen, \IEEEmembership{Senior Member, IEEE}, 
Toshiyo Tamura, \IEEEmembership{Senior Member, IEEE}, 
Naoaki Ono, MD Altaf-Ul-Amin,  Shigehiko Kanaya, Ming Huang$^{\dagger}$, \IEEEmembership{Member, IEEE}
\thanks{
This research and development work was supported by the Grant-in-Aid for Early-Career Scientists \#20K19923.}
\thanks{Zheng Chen is with the Graduate School of Engineering, Osaka University, Japan. 
Ziwei Yang, Lingwei Zhu, Ming Huang, Naoaki Ono, MD Altaf-Ul-Amin, and Shigehiko Kanaya are with the Graduate School of Science and Technology, Nara Institute of Science and Technology, Japan.
Wei Chen is with the Center for Intelligent Medical Electronics, Department of Electronic Engineering, School of Information Science and Technology, Fudan University, Shanghai 200433, China.
Toshiyo Tamura is with Institute for Healthcare Robotics, Waseda University, Japan.
The source code is available \url{https://github.com/zhengchen3/Transformer_Sleep}.
}
\thanks{
$\dagger$Corresponding: Zheng Chen (chen.zheng.bn1@gmail.com);\\\indent \indent Ming Huang (alex-mhuang@is.naist.jp)}
}

\maketitle

\begin{abstract}
Stage-based sleep screening is a widely-used tool in both healthcare and neuroscientific research, as it allows for the accurate assessment of sleep patterns and stages. 
In this paper, we propose a novel framework that is based on authoritative guidance in sleep medicine and is designed to automatically capture the time-frequency characteristics of sleep electroencephalogram (EEG) signals in order to make staging decisions.
Our framework consists of two main phases: a feature extraction process that partitions the input EEG spectrograms into a sequence of time-frequency patches, and a staging phase that searches for correlations between the extracted features and the defining characteristics of sleep stages. 
To model the staging phase, we utilize a Transformer model with an attention-based module, which allows for the extraction of global contextual relevance among time-frequency patches and the use of this relevance for staging decisions.
The proposed method is validated on the large-scale Sleep Heart Health Study dataset and achieves new state-of-the-art results for the wake, N2, and N3 stages, with respective F1 scores of 0.93, 0.88, and 0.87 using only EEG signals. Our method also demonstrates high inter-rater reliability, with a kappa score of 0.80. 
Moreover, we provide visualizations of the correspondence between sleep staging decisions and features extracted by our method, which enhances the interpretability of the proposal. 
Overall, our work represents a significant contribution to the field of automated sleep staging and has important implications for both healthcare and neuroscience research.
\end{abstract}


\section{Introduction}
\label{sec:introduction}

\IEEEPARstart{S}{leep} is an essential human function whose characteristics are manifested by a sequence of physiological alterations, e.g., neural spiking, cardiorespiratory, blood oxygen saturation, and eye activity \cite{PPI}. 
Stage-based sleep screening is currently not only a major tool in the assessment of pathophysiology but also an ingredient in the exploration of neuroscience \cite{Me2, m3,EEGscience}. 
Experimental results have verified some electroencephalogram (EEG) features serve as biomarkers for different sleep stages showing extraordinary physiological significance.
For instance, slow waves contribute to memory consolidation \cite{boosting}, and the sleep spindle is highly correlated to intellectual ability \cite{IQ}. 
Therefore, determining sleep stages and sleep macrostructures is indispensable to healthcare and to facilitating neuroscientific findings.

Formally, the American Academy of Sleep Medicine (AASM) divides sleep into five different stages based on their distinct features. 
There is a constant cycle of sleep stages from wake to non-rapid eye movement (NREM) and REM, where NREM can be further divided into three stages, i.e., N1, N2, and N3 \cite{AASM}.
This cycle repeats several times a night.
Clinically, sleep experts visually inspect the polysomnography (PSG) recordings (especially EEG), and manually score each 30-second signal episode (officially termed \textit{epoch}) to a stage by following the AASM guideline and criteria \cite{PSG}.
This laborious process inevitably limits large-scale applications such as batch staging processes and downstream tasks.
Recent advances in portable monitoring have made possible daily sleep screening with fewer sensors and hence the large volume of sleep data can be collected at a lower cost \cite{ZAx, ZEO}.
However, a reliable automated sleep staging mechanism that could promptly yet accurately score sleep stages is urgent to meet the growth of the sleep community.

Benefiting from the booming of deep learning, various EEG-based frameworks have proven the potential of automated sleep staging for replacing the manual procedure \cite{sp3, sp5, sp6}.
Especially, incorporating the clinical rules of human scoring into powerful deep learning models has become a prevailing trend \cite{chenbibm2021, sleeptrans}. 
For instance, several studies proposed to model the sleep macrostructure such as inter-epoch correlations \cite{featurenet1} or temporal transition \cite{Ziweibibm}, using state-of-the-art deep sequential models, e.g. the Long-Short Term Memory (LSTM) or Transformer \cite{CHENmethods}.
At the core of these studies was to view the staging procedure as sequence-to-sequence classification: the model extracts useful sequential order information for classification from a sequence of input epochs \cite{nn1, sleeptrans}. 

Although the aforementioned approaches could to some extent improve staging performance, they require complicated optimization to adapt to sleep EEG signals to obtain useful representations, which is typically difficult to efficiently perform: EEG features are transient, temporally random, and redundant \cite{chenijcai}.
For example, K-complex (with duration 0.5$\sim$1.5 seconds) is the indicator feature for N2. 
However, K-complex often spontaneously bursts only once in a 30-second N2 epoch, rendering the epoch dominated by massive stage-irrelevant features \cite{wake}.
The above characteristics of EEG signals pose a challenge to conventional methods which often fall short in representation power. 
To meet the nowadays requirement of downstream tasks, 
high-quality EEG stage-dependent features are demanded by properly representing each sleep epoch.

\begin{figure*}[t]
\centering
\includegraphics[width=0.8\linewidth]{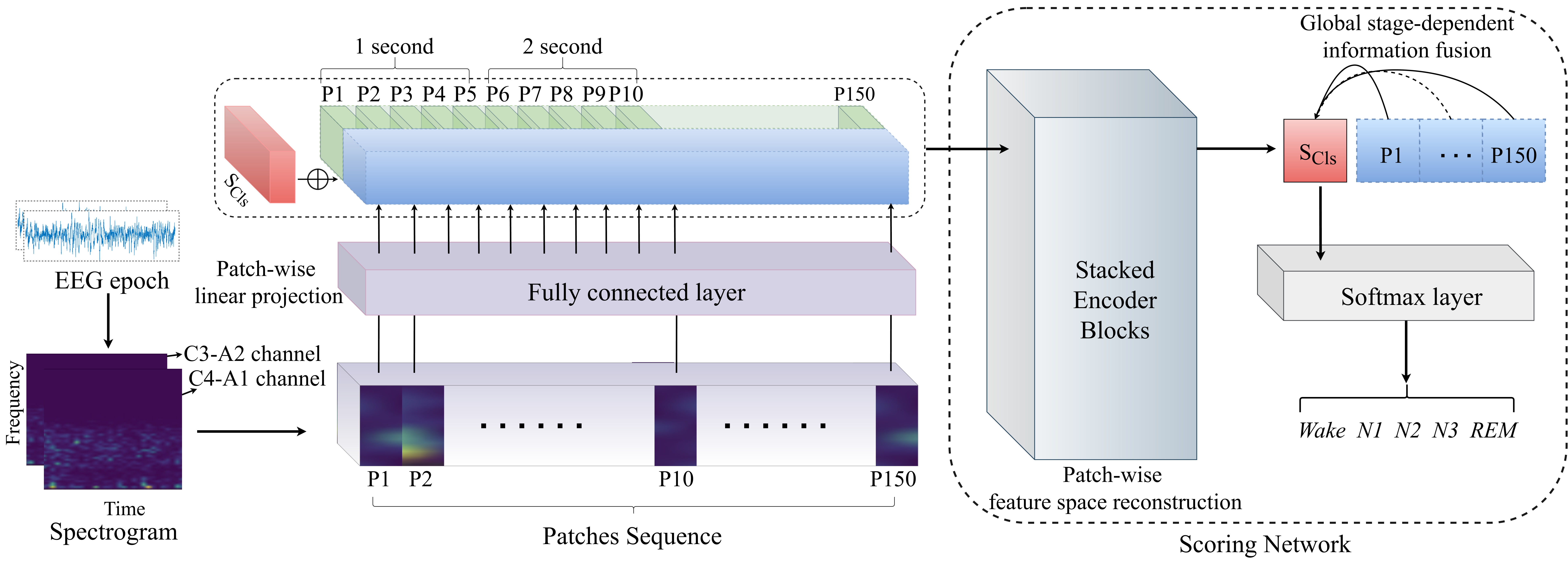}
\caption{System overview. The experiment of this work was done with two-channel EEGs. Each 30-second EEG epoch is firstly transferred to a time-frequency spectrogram (size: $32\times30$). Then, spectrograms of the 2 channels are segmented into patch sequences respectively. Each patch as a 1-second-1-frequency-band feature vector is projected to a high dimension by using a patch-wise fully connected layer. After preprocessing, the augmented patch sequence is fed to the staging model that contains stacked encoder blocks and the final classification layer. After training, a class patch that absorbs intra-patch characteristics is used for stage assignment decisions.}
\label{fig:f2}
\end{figure*}

In this study, we argue that capturing stage-specific features satisfying the clinical criterion plays a vital role in automatic sleep staging. 
This paper proposes a framework designed on top of the EEG-based technical specifications in AASM guidelines that are dedicated to stage-specific feature representation.
This framework is composed of two phases: a feature processing and a staging phase.
The feature processing phase refines an EEG epoch to a feature set containing time series of features from the frequency domain, while
the staging phase inspects this feature set by an elaborate attention-based model, i.e. Transformer, and further distills stage-specific features for staging decision.
The motivation is to retain as high as possible the resolution of frequency while alleviating the influence of EEG signals such as temporal randomness and transiency.
To better discover important features that have strong correlations to stages, we leverage the attention mechanism designed to extract global relevance between time-frequency series.

We further record the relevance when staging an input epoch and visualize the records to present the parts to which our model pays more attention.
Suppose the proposed model is properly trained, then the Transformer is expected to output attentions that could reveal physiologically interpretable patterns important for sleep staging.
To validate the effectiveness of the proposed framework, we experiment with two large-scale benchmark datasets: Sleep Heart Health Study (SHHS) and Sleep-EDF. Our results indicate the proposed framework attains a new state-of-the-art compared to various existing studies.
We summarize the main contributions as the following: 
\begin{itemize}
 \item This work proposes a staging framework for identifying stage-specific features from EEG data in accordance with the sleep community's definition of these characteristics. Taking the clinical rules of scoring stages into account, the framework is capable of improving staging accuracy and providing intuitive explanations.
\item The proposal achieves the state-of-the-art (SOTA) classification performance by the novel network structure with only EEG signals.
\item We propose a relevance-based method to visualize the resultant attention matrix and therefore equip the model with strong interpretability for its staging decisions.
\end{itemize}

The remainder of this paper is organized as follows. 
Section \ref{sec:related} discusses previous work and describes stage-specific definitions in the AASM guidelines. 
Section \ref{sec:method} introduces the proposed framework. 
Section \ref{sec:experiment} presents the experiment setup followed by results in Section \ref{sec:result}. 
We discuss the results in Section \ref{sec:discussion} and finally conclude the paper in Section \ref{sec:conclusion}.

\section{Related Work and Preliminary}
\label{sec:related}
\subsection{Related Work}

Facilitating efficiency in staging is a long-standing problem in the sleep community. 
Staging in automation is an often sought-after solution fueled by novel deep learning techniques. 

Building on top of deep learning models, it is vital for any downstream tasks to properly represent the data or the input of the model.
Conventionally such representation extraction is done by manually selecting/crafting features from statistical methods \cite{sp1,chensybolic,stas1}, power spectral density \cite{psd1,psd2specenstas,psd3}, information entropy \cite{entro}, wavelet transformation \cite{cw1,reviewer2_1,revier2_2}, empirical mode decomposition \cite{emd1}, etc.
However, extracting hand-crafted features is laborious and requires heavy prior knowledge.
Practitioners often select features that are expected to work well with the model from a statistical learning perspective, and seldom consider the feature extraction from a neuroscientific perspective that explicitly considers the intrinsic nature of EEGs \cite{EEGeverchange}.

It has been demonstrated that incorporating automatic feature extraction using feature-mapping-oriented could improve staging performance by a large margin \cite{psd2specenstas,CHEN}.
Those works typically leveraged convolutional neural networks (CNNs) to capture more informative frequency features from EEG recordings \cite{nn2, nn3,Enrique}.
Though the convolution operation can express temporal connectivity of sleep rhythms within a neighborhood, the shift-invariant nature of CNNs leads to the loss of information in the global context, e,g., transitional information in sleep.
A reasonable substitution is to employ the inherently temporal-based models, i.e., recurrent neural networks (RNNs) \cite{sp4, ShreyasPATHAK,shuiwang}.
Different from the CNNs, RNN-based models pay attention to the information of global context by allowing sequential modeling of dependency and transfer of temporal influence \cite{nornn,wake}.
However, the global perspective and the inherently sequential nature (time-invariant) of RNNs conflict with the instantaneous sensitivity of EEGs, which leads to the loss of feature extraction quality~\cite{chenijcai}. 

Recent work attempts to introduce the Transformer \cite{attenisalluned} to sleep staging and relevant tasks \cite{sleeptrans,atten1,atten2}.
Qu et al. adopt residual blocks after Hilbert-Transform-like preprocessing and show an accurate stage performance by the Transformer \cite{nn1}.
Phan et al. extend their work \cite{sp4} to an entirely Transformer-based framework, and the performance is shown to significantly improve over prior methods \cite{sleeptrans}.
However, all aforementioned existing work still applied the sequence-to-sequence strategy such as CNNs, RNNs, Transformers, or combinations of them. 
Hence their respective drawbacks still persist.
Figuring out a configuration that can get rid of those shortcomings is nontrivial.



\subsection{Preliminary: Notes on EEG Representation}
Quantitative analysis in EEG has revealed that there are electrical rhythms associated with different sleep stages \cite{EEGscience}.
The AASM guideline defines the staging rule with the range 0.5 to 30-35 Hz in the frequency oscillation \cite{AASM}. 
Specifically, wakefulness can be identified by a dominant ($>$50\%) alpha rhythm (8-13 Hz), and associated beta waves (16-32 Hz) \cite{wake}.
Theta waves (4-8 Hz) that consist of low-amplitude and mixed frequency waves are the indicator of stage N1. 
When one or more K-complex and sleep splines under low beta waves (12-16 Hz) appear, the corresponding epoch will be scored as stage N2, while slow delta waves (1-4 Hz) in EEG are the dominant feature of stage N3. 
Although the predominant characteristic of REM is the movements in the eye and muscle, sawtooth waves in EEG recording are often considered as the alternative.
In this paper, we aim to build a framework that embodies the above-mentioned rules in its automated staging, and a system overview of the proposed framework has shown in Fig. \ref{fig:f2}.

\begin{figure}[t]
\centering
\includegraphics[width=0.98\linewidth]{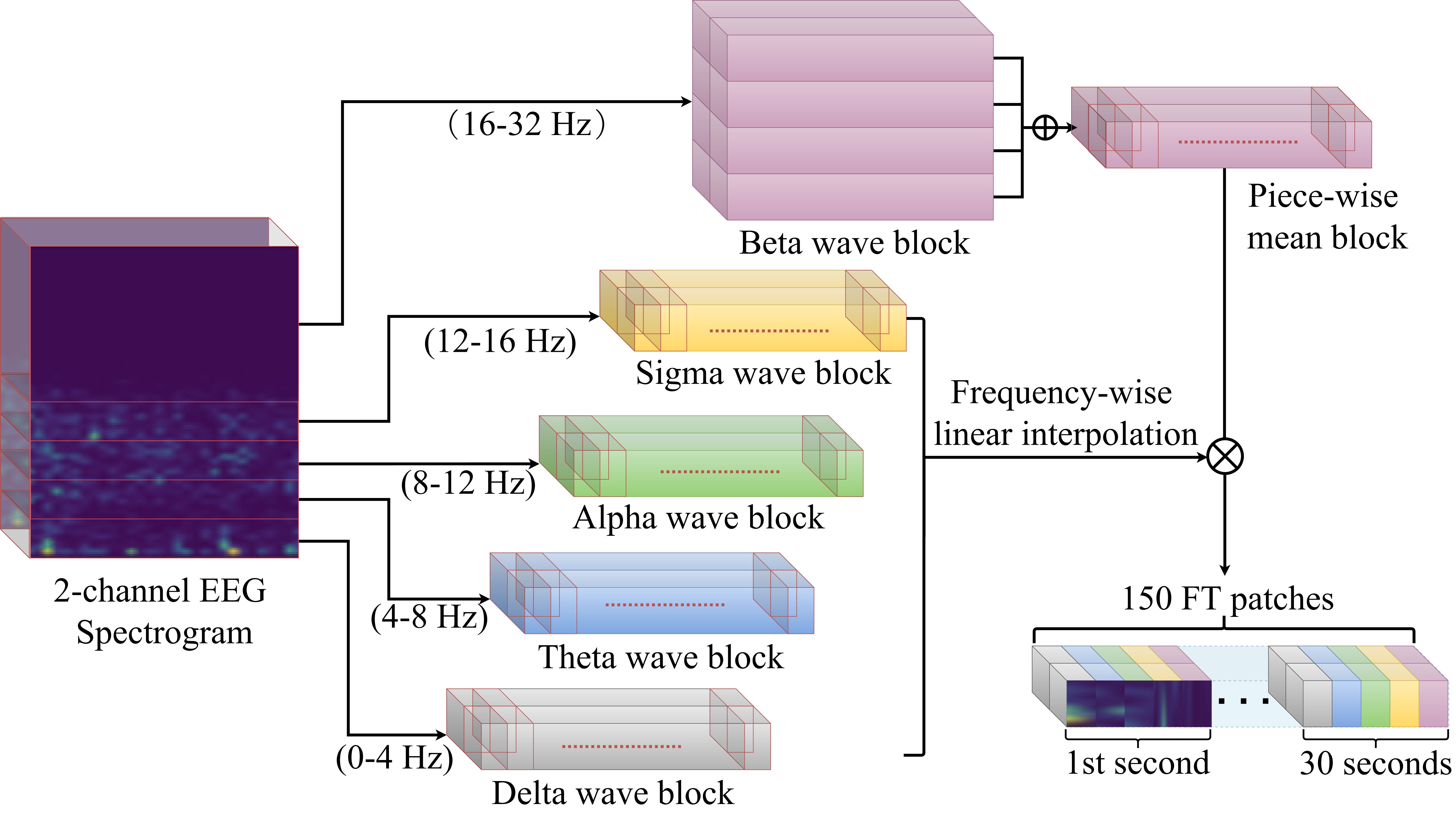}
\caption{Workflow of time-frequency patching.}
\label{fig:f3}
\end{figure}


\section{Methods}
\label{sec:method}


\subsection{Time-Frequency Representation}
\label{Time-Frequency}
EEG recordings typically are contaminated with various types of noises, we first apply an 8th-order Butterworth band-pass filter with a 0.2-32 Hz cut-off band to each recording. 
Then, each 30-second epoch is transformed into a spectrogram to represent the time-frequency feature. 
Considering the transient sleep rhythm is continuously burst within a 1-second duration, we generate a log-power spectrogram from the EEG signal for every second by using a non-overlapping Hamming window and fast Fourier transformation.
Meanwhile, an integral in the power spectrum is calculated every 1 Hz, and this spectrogram hence provides distinguishable features in time-frequency resolution for different sleep stages.

Experimentally, two spectrograms corresponding to two EEG channels are generated for each epoch. 
Let us denote a spectrogram as $S\in\mathbb{R} ^{F\times T\times C}$, where $F$ denotes the frequency range (0-32 Hz), $T$ denotes the time (30-seconds), and $C$ denotes the number of channels.

\subsection{Frequency-time Patching}
\label{Generation}

Building on the foundation of EEG data pre-processing, we propose a feature processing framework termed \emph{frequency-time} (FT) \emph{patching} to refine the EEG feature representation.
Each spectrogram is first divided into eight frequency bands with 4 Hz: this setting corresponds to the standard four frequency bands, i.e., Delta ($\delta$), Theta ($\theta$), Alpha ($\alpha$), Sigma ($\sigma$) and four sub-bands derived from further dividing the Beta $\beta$ band (\emph{frequency patching}).  
Subsequently, time patches are acquired by extracting and rearranging each 1-second column of the spectrogram (\emph{time patching}). 
This framework is on top of two important observations: a single epoch may contain multiple stages \cite{region} while the bursting time of almost clinically crucial features is within the 1-second resolution \cite{trasientissue}. 
Moreover, we propose a special network architecture detailed in Section \ref{Network} corresponding to the \emph{frequency-time patching}  that could provide model interpretability. 
The workflow (seen in Fig. \ref{fig:f3}) of the \emph{frequency-time patching} is as follows:

\begin{itemize}
\item \emph{Frequency patching}: a spectrogram $S$ is split into five parts $S = (S_{\delta},S_{\theta},S_{\alpha},S_{\sigma},S_{\beta})$ in accordance with the five predominant frequency bands  of sleep rhythms (see Appendix TABLE V). 
The first four components, $S_{\delta}$, $S_{\theta}$, $S_{\alpha}$, $S_{\sigma}\in\mathbb{R} ^{(F/8)\times T\times C}$ have the same bandwidth of 4 Hz. 
The beta band is further divided into four sub-blocks, 
$S_{\beta}$ = ($S_{\beta1}$, $S_{\beta2}$, $S_{\beta3}$, $S_{\beta4}$), where the subscripts correspond to the four quarters of the beta band. 
\item \emph{Time patching}: Each frequency block is divided by column to extract frequency-time patches (middle of Fig. \ref{fig:f3}). 
We denote every time patch as $S^{i}=(S^{i}_{\delta},S^{i}_{\theta},S^{i}_{\alpha},S^{i}_{\sigma},S^{i}_{\beta}$, with the superscript $i$ indicating the \emph{i-th} second. 
\item \emph{Sub-block averaging}: We compute the mean values of the sub-blocks of the beta band $\bar{S}_{\beta}$ = ($\bar{S}_{\beta1}$, $\bar{S}_{\beta2 }$, $\bar{S}_{\beta3}$, $\bar{S}_{\beta4}$).
\item \emph{Spectrogram transformation}: We build a new structure $S'^{i}=(S^{i}_{\delta},S^{i}_{\theta},S^{i}_{\alpha},S^{i}_{\sigma},\bar{S}^{i}_{\beta})\in\mathbb{R}^{20}$, $S'\in\mathbb{R} ^{20\times T\times C}$.
\item \emph{Patches rearrangement}: A sequence of frequency-time patches $S_{seq}$ is generated by column-wise traversal. 
$S_{seq} \in \mathbb{R}^{150 \times D_{p}}$ ($150 = 5 $ frequency bands $\times 30$ second) is the patch sequence for one epoch, $D_{p}$ denotes the dimension of each patch. 
\end{itemize}

We validate our feature processing framework against various data processing ablations in Section \ref{sec:baseline}.

\begin{figure}[t]
\centering
\includegraphics[width=0.98\linewidth]{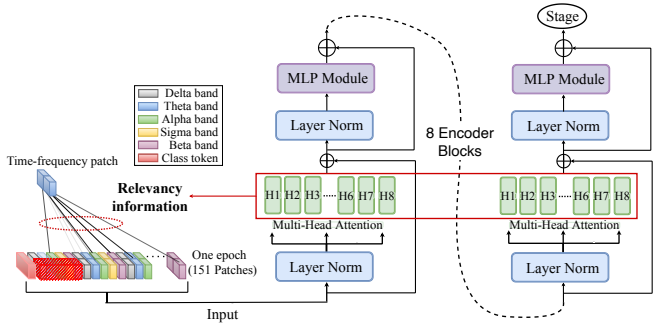}
\caption{Architecture of the stacked Transformer encoder blocks.}
\label{fig:fig4}
\end{figure}

\subsection{{Patches Sequence Embedding}}
\label{Embedding}

The individual FT patches contain uneven information about different sleep stages while the concurrence of patches and their temporal order are also informative.
Therefore, we propose to exploit the self-attention mechanism popular in the current deep learning community \cite{ViT1} to extract such hidden information from the concurrence in the patches sequence.
Specifically, the Transformer model is utilized as the backbone of the staging framework, which individually summarizes the relevance of patches called for.
The Transformer creates a set of sub-networks to handle different feature sub-spaces while relevant features among these sub-networks are subsequently found by the embedded attention mechanism \cite{chenbibm2021}.
The Transformer naturally suits our framework for processing and tracking individual contributions of the FT patches. 

Algorithmically, the Transformer does not explicitly generate a new feature tensor for downstream tasks.
Inspired by \cite{BERT}, an extra parameterized patch $S_{Cls}$ is created and appended at the beginning of the patch sequence.
Accompanying the data passing through the model architecture, this patch is retained as the stage indicator summarized the global relevance of FT independence that can serve for the final staging step.
This process is formulated as below:
\begin{equation}
{S}'_{seq} = \texttt{Concat}(S_{Cls}, \,E\cdot S_{seq}),
\end{equation}
where $E$ is a patch-wise linear projection that enriches $S_{seq}$ to a higher informative dimensional space. 
${S}'_{seq} \in\mathbb{R} ^{(150 +1)\times D}$ is the output sequence where $D$ is the output dimension  of the linear projection.

Unlikely the RNNs, the Transformer leaves out the sequential order or positional information of patches, 
therefore, we make use of the positional embedding technique \cite{ViT1} to merge a sequence of learnable positional patches ($E^{pos}_{\phi}$) into the ${S}'_{seq}$ as the final input:
\begin{equation}
X={S}'_{seq} + E^{pos}_{\phi},
\end{equation}
where $E^{pos}_{\phi}\in\mathbb{R} ^{(150+1)\times D}$ has the same shape as the ${S}'_{seq}$, and $\phi$ illuminates the learnt weights of the positional patches.



\begin{figure}[t]
\centering
\includegraphics[width=0.95\linewidth]{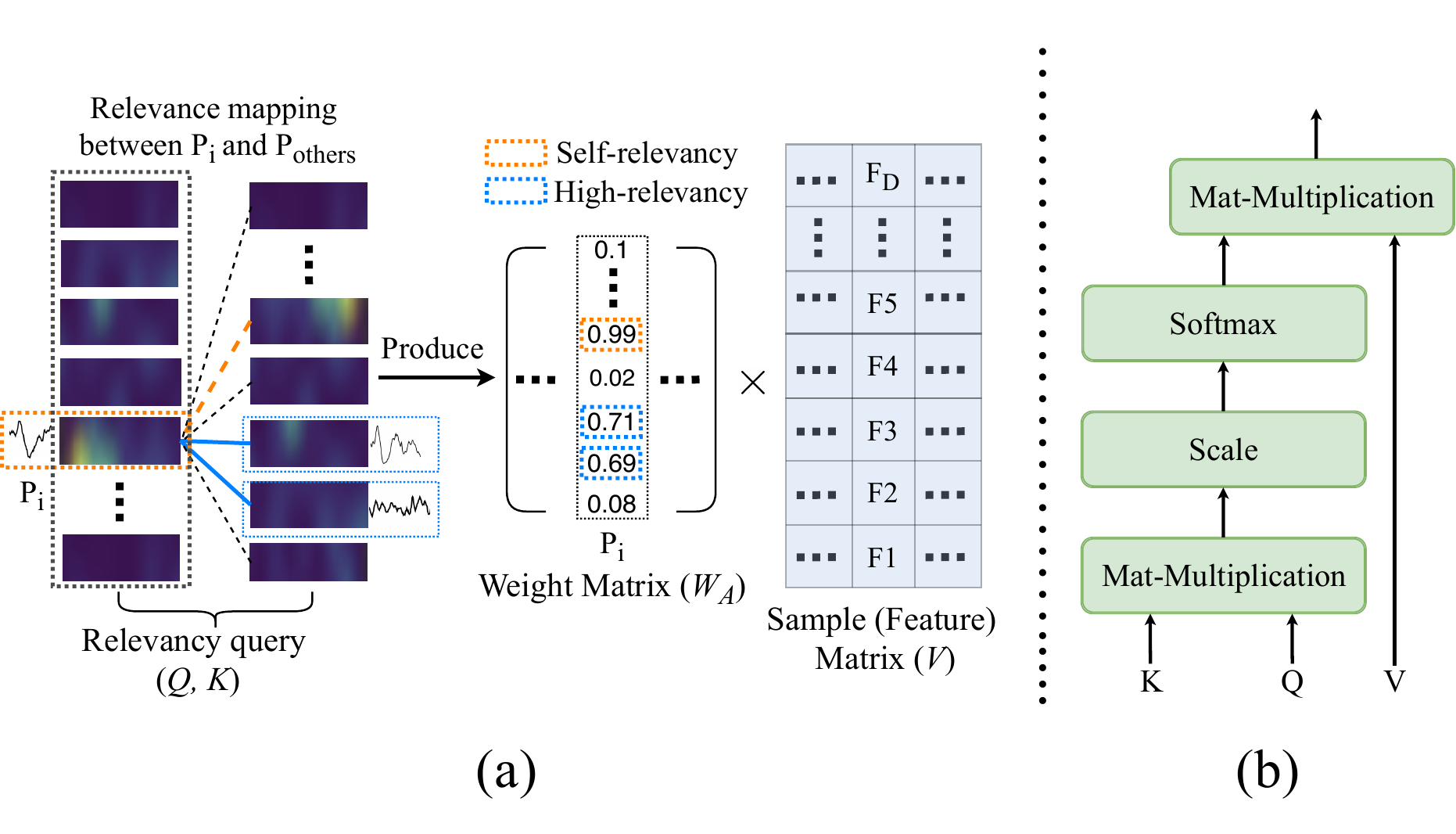}
\caption{Attention mechanism: (a) shows that the attention layer first calculates relevancies among patches and then maps the relevant weight matrix to an input of each attention layer. (b) illuminates the workflow of the attention layer.}
\label{fig:atten}
\end{figure}

\subsection{{Transformer Architecture}}
\label{Network}
The Transformer architecture in our proposal is composed of a multi-layer perceptron (MLP) followed by an attention layer.
Meanwhile, a LayerNorm operation $ \texttt{LayerNorm}(\cdot)$ is inserted at the beginning of each functional layer and a residual connection functions after each LayerNorm as shown in Fig. \ref{fig:fig4}. 
The attention layer calculates the pair-wise relevance of different patches in $X$.
The Patch-wise MLP comes from the fact that information propagation is restricted to the same patch and renders the network to calculate the relevance of patches by self-attention only. 


\noindent\textbf{Attention mechanism:}
The attention mechanism assigns a relevance score of the ground-truth stage to each input patch. 
It comprises three components: query ($Q$), key ($K$), and value ($V$) matrices, which are the linear projections of the input $X$.
The matrix $Q$ represents a query sequence with FT patches. 
In the case of self-attention, $K$ is identical to $Q$, hence the attention utilizes a matrix multiplication within $K$ and $Q$ to calculated relevance among the patches since each row or column in the matrix can be regarded as a projected space of one patch. 
The resultant matrix $A_{patch}$ records the attention score in different patches by weighting the relevance resulted in $W_{A}$ to each row of $V$.
The above description has shown in Fig. \ref{fig:atten} and the mathematical process can be summarized as:
\begin{equation}
\label{eq7}
A_{patch} = W_{A}\cdot V, \indent \text{where   } W_{A}=\sigma(\frac{QK^{T}}{\sqrt{d}}).
\end{equation}
where $\sigma(\cdot)$ denotes the softmax operation,
$\sqrt{d}$ is a normalization operation that is applied to each $Q$-$K$ computation.
Moreover, each attention layer simultaneously outputs $h$ attention matrices to map the diversity of the outputs, commonly called "heads" \cite{CHENmethods}.
This implementation is called "multi-head attention" and is utilized in this paper.

\begin{figure}[t]
\centering
\includegraphics[width=0.92\linewidth]{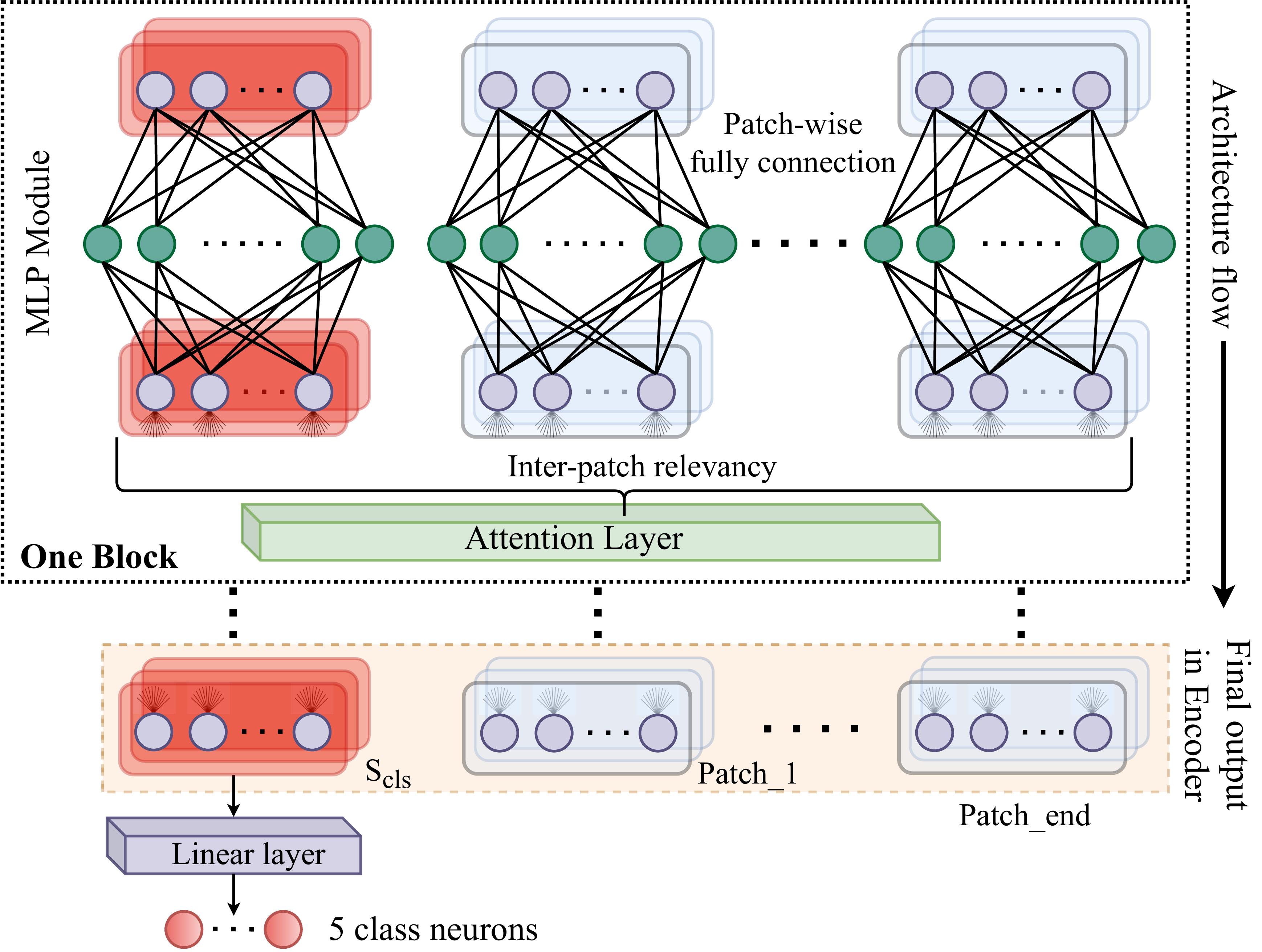}
\caption{Patch-wise MLP architecture and final staging module}
\label{fig:MLP}
\end{figure}


\noindent\textbf{Patch-wise MLP \& Staging module:}
The normalized output of the multi-head attention layer is then fed to the MLP layer. 
The MLP contains two linear layers with Gaussian error linear units (GELUs) and a residual connection aiming to avoid the gradient vanishing issue. 
As the information passes through all stacked blocks, the class patch ${S}'_{Cls}$ has absorbed information about the relevance of FT patches from the global context and is used solely for the staging decision. 
As shown in Fig. \ref{fig:MLP}, a linear projection finally compresses the flattened class token ${S}'_{Cls}$ to the same number of neurons to the sleep stages. 
\begin{equation}
y = \texttt{Linear}(\,\texttt{LayerNorm}(\,{S}'_{Cls}\,)\,).
\label{eq11}
\end{equation}
where $y \in$ \{W, N1, N2, N3, REM\}, and ${S}'_{Cls}$ is also normalized before the final classification (linear) layer.

\subsection{{Attention Visualization}}
\label{av}
As outlined above, attention computation relies heavily on matrix multiplication and the attention scores play different roles in staging.
To provide interpretability of the Transformer and reveal to what extent these stage-specific patches affect the decision-making, we use an attention-oriented visualization \cite{visual} to retrieval these FT patches to which the Transformer pays the most attention.
This attention/importance calculation is in terms of both the gradient and relevance information starting from the final classification decision to each attention layer. 
Here, the visualization output a reconstructed spectrogram-like graph $\hat{V}$ rendered by importance calculations of patches and the definition is as follow:
\begin{equation}
\hat{V} = \bar{A}^{(1)}\odot \bar{A}^{(2)} \odot \dotsc \odot \bar{A}^{(B)},
\end{equation}
where $\odot$ is the Hadamard product and
$\bar{A}^{(i)} \in\mathbb{R} ^{{F}'\times T} , i\in[1,\dots, B]$ is the index of $B$ Transformer encoders. 
Since the rows of $W_{A}$ in Eq. \ref{eq7} are normalized, $W_{A}$ can be treated as a type of attention map. 
Each sub-graph $\bar{A}^{(b)}$ of encoder ($b$) records gradient information of the attention map $\bigtriangledown W_{A}^{(b)}$ and its relevance diffusion/propagation $R^{(n_{b})}$, that is,
\begin{equation}
\bar{A}^{(b)}=I + \mathbb{E}_{h}\big[\bigtriangledown W_{A}^{(b)}\odot R^{(n_{b})}\big],
\end{equation}
where $\mathbb{E}_{h}$ denotes the expectation w.r.t. the multi-head and an identity matrix $I$ prevent self-inhibition issue in patches \cite{visual}.

The relevance propagation $R^{(n}$ starts from the classification layer in the staging module and then iteratively diffuses to each previous layer $L^{(n)}, n\in(1, \cdots N)$ until the Transformer input,
hence the classification layer is defined as $L^{(1)}$. 
Let $L^{(n)}(X^{(n)},W^{(n)})$ denote the $n$-th layer on its input $X^{(n)}$ and weights $W^{(n)}$, the relevance propagation is similar to the chain rule that the derivative of each attention map referring to the generic Deep Taylor Decomposition \cite{tyler}:
\begin{equation}
R^{(n)}_{j} = \sum_{i}^{}X^{(n)}_{j}\frac{\partial L^{(n)}_{i}(X^{(n)},W^{(n)})}{\partial X_{j}}\frac{R^{(n-1)}_{i}}{L^{(n)}_{i}(X^{(n)},W^{(n)})},
\end{equation}
where the subscripts $i, j$ are patch indices. 
Since two-channel EEGs are used in this work, each input generates two attention graphs. 

\begin{table}[t]
\centering
\caption{Dataset description of SHHS, Sleep-EDF, and 'health-set'}
\label{tb:dataset description}
\begin{tabular}{c|cc|c} 
\toprule
\multirow{2}{*}{}   & \multicolumn{2}{c|}{SHHS}                          & Sleep-EDF      \\ 
\midrule{}
                        & All              & Healthy-set                     & All                   \\ 
\midrule
\#Subject               & 5736             & 684                             & 183                            \\
\multirow{2}{*}{Gender} & M: 2774          & M: 360                          & M: 74          \\
                        & F: 2962          & F: 324                          & F: 109                      \\
Age               & 62.17$\pm$11.02  & 63.14$\pm$11.22                 & 18-102                  \\ 
\midrule
\#W                     & 1666191 (28.8\%) & \multirow{5}{*}{~26080 (20\%)~} & 40638 (26.3\%) 
  \\
\#N1                    & 214985 (3.7\%)   &                                 & 18013 (11.7\%)                     \\
\#N2                    & 2371496 (40.9\%) &                                 & 52341 (39.8\%)                   \\
\#N3                    & 732389 (12.6\%)  &                                 & 11467 (7.6\%)                    \\
\#REM                   & 809155 (14.0\%)  &                                 & 22304 (14.5\%)                     \\
\bottomrule
\end{tabular}
\end{table}

\section{{Experiment}}\label{sec:experiment}

\subsection{{Dataset and Preprocessing}}

\textbf{SHHS dataset:}
Sleep Heart Health Study (SHHS) is a large-scale sleep database devoted to investigating whether sleep-related breathing is associated with an increased risk of some diseases.
The SHHS contains two rounds of at-home PSG recordings.
The first round (SHHS-1) consisting of 5793 individuals is used in this work. 
Sleep stages were scored by consensus  between two sleep technicians who were blind to the condition of the participants for six classes (wake, REM, S1, S2, S3, S4) according to the R\&K guidelines \cite{KRrule}, this work merges S3 and S4 into stage N3 by referring to the AASM criteria \cite{AASM}.

Considering the imbalance issue (refer to TABLE \ref{tb:dataset description}) of SHHS dataset, we create a balanced subset called \textbf{\emph{healthy-set}} that serves in the pre-training phase in our experiments.
The candidate selection in this \emph{healthy-set} is based on six clinical indicators since SHHS-1 provides a personal health description of all subjects, and the detailed description can be seen in Appendix VII-A.
As a result, the healthy-set consists of 684 subject-wise recordings with 26080 epochs for each class.

\textbf{Sleep-EDF expanded dataset} used in our experiments is an expanded version to verify the generalization of the proposal. 
It consists of 197 whole-night PSG recordings from two types of sleep files: Sleep Cassette and Sleep Telemetry file.
All subjects experimented with in this database are aged between 18-101. 
The recordings contain other two channels of EEG signals instead of C4-A1/C3-A2 channels, i.e., Fpz-Cz and Pz-Oz with a 100-Hz sampling rate. Due to detection failure, 183 EEG recordings were ultimately chosen in this paper. 
A summary of this dataset and a detailed description of healthy-set can be found in Table \ref{tb:dataset description}.

\subsection{Training Strategy}
To tackle the issue of data unbalances, the experiments are implemented on a pre-training-to-fine-tuning strategy.
Specifically, during the pre-training phase, the healthy-set is used to initialize the model parameters. 
The $AdamW$ optimizer with a big learning rate of $10^{-3}$ is utilized for the optimization.
In the fine-tuning phase, we utilize the remaining data to exhaustively train the pre-trained model.
Also, the $AdamW$ optimizer with $10^{-4}$ learning rate is used to meticulously optimize the classification loss (i.e., cross-entropy loss function) \cite{Adamw}.

To alleviate the overvaluation of the performance, we implement a subject-wise 7-fold cross-validation by splitting the data into seven subject-wise subsets. 
In each trial, six subsets are used in the training step, while the remaining subset (roughly 800 subjects) is used for validation.

\begin{table}[t]
\caption{Parameter settings of the proposed method and the optimal combination is in bold.}
\centering
\begin{tabular}{l|c}
\toprule
\textbf{Parameter}                        & \textbf{Value}   \\ \midrule
\#Stacked encoder                         & \{6, \textbf{8}, 12\} \\
\#Heads ($h$)                         & \{2, 4, \textbf{8}, 12\}      \\
Dimension of linear projection of $D$ & \{16, \textbf{32}, 64\}   \\
Normalization-like scale ($\sqrt{d}$)               & \{2, \textbf{4}\}        \\
Dimension of MLP output                          & \{64, \textbf{128}, 256\} \\
Dropout rate                              & \{0.2, \textbf{0.5}, 0.8\} \\
\#Training epoch                              & 200 \\
Batch size                              & 32\\
\#Parameters                              & $1.3 \times 10^{5}$\\
\bottomrule
\end{tabular}
\label{tb3}
\end{table}

\subsection{Parameter Settings}
The details of the parameter settings have shown in TABLE \ref{tb3}.  
Here, we implement a grid search to find the best combination of parameters, and the optimal settings (the bold values in TABLE \ref{tb3}) are used both for pre-training and subsequent training. 
Additionally, a \emph{dropout} operation is added after each linear projection and attention layer to further avoid overfitting. 
All experiments are conducted on a server with the NVIDIA GeForce RTX 3090Ti GPUs. 

\subsection{Metrics and Comparisons}\label{sec:baseline}
Three metrics are used in the experiments to evaluate the staging performance, i.e., the stage-specific precision ($Pre$), recall ($Re$), F1-score ($F1$), overall accuracy ($Acc$), and Cohen's Kappa coefficient ($k$) to measure the inter-rater reliability \cite{kappa}.

We comprehensively design different ablation studies from three factors to evaluate the effectiveness of our proposal: data processing, model architecture, and learning strategy.

\noindent\textbf{Data processing} ablation studies are conducted to demonstrate how the proposed framework performs when removing \emph{FT patching} feature processing against:
\begin{enumerate}
    \item time-domain patching spectrogram that we  use 1-second time patches without frequency patching to prove the necessity of frequency refinement;
    \item multi-scale EEG-oriented CNNs (termed by \emph{Inception}) proposed in work \cite{incep}. Concretely, a convolutional layer containing five filters corresponding to the five frequency bands was constructed. The output can be viewed as the filter-based refinement in different frequency bands.
\end{enumerate}
\noindent\textbf{Sequential model} ablation is enlightening to study how the performance varies by replacing the Transformer since attention calculation plays the central role of extracting representational information for the staging decision.
We compare with:
\begin{enumerate}
\addtocounter{enumi}{2}
    \item Time patching+LSTM; \space\space\space\space\space\space\space\space\space 5) FT + LSTM; 
    \item Time patching+Bi-LSTM; \space\space\space\space\space 6) FT + Bi-LSTM. 
\end{enumerate}
\noindent\textbf{Learning strategy.} Although the model is pre-trained with the balanced \emph{healthy-set}, the imbalance data issue persisted in the fine-tuning phase. 
Inspired by previous work \cite{nn1} which tried to overcome the imbalance problem with a weighted loss function, we implemented a class-wise weighted cross-entropy loss function \cite{WCEunet} as the ablation (7).

\section{{Results}}\label{sec:result}

\begin{table*}[t]
\centering
\caption{Performance obtained by proposed method and existing works using SHHS and Sleep-EDF database.}
\label{sota}
\begin{tabular}{llcccccccccc} 
\toprule
Dataset                    & System                                               & Method                                                                                    & \#Record &               & Wake          & N1            & N2            & N3            & REM           & $k$    & \emph{Acc}                       \\ 
\midrule
\multirow{24}{*}{SHHS}      &                                                      & \multirow{3}{*}{\begin{tabular}[c]{@{}c@{}}EEG \\+ proposed method\end{tabular}}                                                    &                                                           & \textit{Pre}  & 0.93          & 0.42          & 0.87          & \textbf{0.89} & 0.80          & \multirow{3}{*}{0.80}      & \multirow{3}{*}{0.85}      \\
                            & \textbf{Proposal}                           &                                                                                           & 5736                                                      & \textit{Re}   & \textbf{0.93} & 0.33          & \textbf{0.90} & 0.84          & 0.79          &                                 \\
                            &                                                      &                                                                                           &                                                           & \textit{F1}   & \textbf{0.93} & 0.38          & 0.88          & \textbf{0.87} & 0.80          &                                 \\ 
\cline{2-12}
                            &                                                      & \multirow{3}{*}{\begin{tabular}[c]{@{}c@{}}EEG sequence \\+ proposed method\end{tabular}} &                                                           & \textit{Pre}  & \textbf{0.94} & 0.50         & \textbf{0.89} & \textbf{0.89} & \textbf{0.88} & \multirow{3}{*}{\textbf{0.85}}     & \multirow{3}{*}{\textbf{0.89}}      \\
                            & \textbf{Proposal + \cite{sleeptrans}}                           &                                                                                           & 5736                                                      & \textit{Re}   & \textbf{0.93} & 0.44          & \textbf{0.90} & 0.85         & 0.85          &                                 \\
                            &                                                      &                                                                                           &                                                           & \textit{F1}   & \textbf{0.93} & 0.47          & \textbf{0.89} & \textbf{0.87} & 0.86          &                                 \\ 
\cline{2-12}
                            &                                                      & \multirow{3}{*}{EEG + Transformer}                                                        &                                                           & \textit{Pre}  & -             & -             & -             & -             & -             & \multirow{3}{*}{0.83}       & \multirow{3}{*}{0.88}    \\
                            & Phan et al., 2022 \cite{sleeptrans}                                &                                                                                           & 5791                                                      & \textit{Re}   & -             & -             & -             & -             & -             &                                 \\
                            &                                                      &                                                                                           &                                                           & \textit{F1}   & 0.92          & 0.46 & 0.88          & 0.85          & 0.88          &                                 \\ 
\cline{2-12}
                            &                                                      & \multirow{3}{*}{\begin{tabular}[c]{@{}c@{}}EEG, EOG, EMG\\+ Separable CNN\end{tabular}}          &                                                           & \textit{Pre}  & 0.89          & \textbf{0.57} & 0.85          & 0.88          & 0.83          & \multirow{3}{*}{0.80}      & \multirow{3}{*}{0.85}     \\
                            & Fernandez et al., 2021 \cite{Enrique}                       &                                                                                           & 5793                                                      & \textit{Re}   & \textbf{0.93} & 0.23          & 0.89          & 0.77          & 0.85          &                                 \\
                            &                                                      &                                                                                           &                                                           & \textit{F1}   & 0.91          & 0.40          & 0.87          & 0.83          & 0.84          &                                 \\ 
\cline{2-12}
                            &                                                      & \multirow{3}{*}{\begin{tabular}[c]{@{}c@{}}EMG, EOG, EEG\\+ GRU, LSTM\end{tabular}}       &                                                           & \textit{Pre}  & -             & -             & -             & -             & -             & \multirow{3}{*}{\textbf{0.85}}  & \multirow{3}{*}{\textbf{0.89}}\\
                            & Phan et al., 2021 \cite{shuiwang}                                &                                                                                           & 5791                                                      & \textit{Re}   & -             & -             & -             & -             & -             &                                 \\
                            &                                                      &                                                                                           &                                                           & \textit{F1}   & 0.92          & \textbf{0.50}          & 0.88          & 0.85          & 0.88          &                                 \\ 
\cline{2-12}
                            &                                                      & \multirow{3}{*}{EEG + CNN}                                                                &                                                           & \textit{Pre}  & 0.90          & 0.30          & 0.87          & 0.87          & 0.80          & \multirow{3}{*}{0.81}    & \multirow{3}{*}{0.85}       \\
                            & Eldele et al., 2021 \cite{emadeldeen}                       &                                                                                           & 329                                                       & \textit{Re}   & 0.83          & 0.36          & 0.86          & \textbf{0.87} & 0.83          &                                 \\
                            &                                                      &                                                                                           &                                                           & \textit{F1}   & 0.86          & 0.33          & 0.87          & \textbf{0.87}          & 0.82          &                                 \\ 
\cline{2-12}
                            &                                                      & \multirow{3}{*}{\begin{tabular}[c]{@{}c@{}}EEG, EOG, EMG\\+ CNN, bi-LSTM\end{tabular}}    &                                                           & \textit{Pre}  & 0.92          & 0.31          & 0.83          & 0.84          & \textbf{0.88} & \multirow{3}{*}{0.79}       & \multirow{3}{*}{0.84}    \\
                            & Pathak et al., 2021 \cite{ShreyasPATHAK}                         &                                                                                           & 5793                                                      & \textit{Re }  & 0.92          & \textbf{0.50} & 0.84          & 0.67          & \textbf{0.89} &                                 \\
                            &                                                      &                                                                                           &                                                           & \textit{F1}   & 0.92          & 0.40          & 0.84          & 0.76          & \textbf{0.89} &                                 \\ 
\cline{2-12}
                            &                                                      & \multirow{3}{*}{EEG + RCNN}                                                               &                                                           & \textit{Pre } & 0.92          & 0.42          & 0.85          & 0.85          & 0.87          & \multirow{3}{*}{0.80}      & \multirow{3}{*}{0.85}     \\
                            & Seo et al., 2020 \cite{SEO}                              &                                                                                           & 5791                                                      & \textit{Re }  & 0.88          & 0.47          & \textbf{0.90} & 0.86          & 0.86          &                                 \\
                            &                                                      &                                                                                           &                                                           & \textit{F1 }  & 0.90          & 0.45          & 0.87          & 0.85          & 0.86          &                                 \\ 

\midrule

\multirow{15}{*}{Sleep-EDF} & \multirow{3}{*}{\textbf{\textbf{Proposal}}} & \multirow{3}{*}{\begin{tabular}[c]{@{}c@{}}EEG \\+ proposed method\end{tabular}}                                              & \multirow{3}{*}{183}                                      & \textit{Pre}  & 0.91          & 0.41          & 0.85          & 0.83 & 0.84          & \multirow{3}{*}{0.79}      & \multirow{3}{*}{0.84}     \\
                            &                                                      &                                                                                           &                                                           & \textit{Re}   & \textbf{0.93} & 0.33          & 0.84 & 0.84          & 0.83          &                                 \\
                            &                                                      &                                                                                           &                                                           & \textit{F1}   & 0.92          & 0.37          & 0.84 & 0.83 & 0.83          &                                 \\ 
\cline{2-12}
& \multirow{3}{*}{\textbf{\textbf{Proposal + \cite{sleeptrans}}}} & \multirow{3}{*}{\begin{tabular}[c]{@{}c@{}}EEG sequence \\+ proposed method\end{tabular}}                                           & \multirow{3}{*}{183}                                      & \textit{Pre}  & \textbf{0.95}          & \textbf{0.49}          & 0.88          & 0.85 & \textbf{0.86}          & \multirow{3}{*}{\textbf{0.81}}         & \multirow{3}{*}{\textbf{0.86}}  \\
                            &                                                      &                                                                                           &                                                           & \textit{Re}   & 0.93 & \textbf{0.51}          & \textbf{0.89} & \textbf{0.86}          & 0.84          &                                 \\
                            &                                                      &                                                                                           &                                                           & \textit{F1}   & \textbf{0.94}          & \textbf{0.50}          & \textbf{0.88} & 0.85 & \textbf{0.85}          &                                 \\ 
\cline{2-12}
                            & \multirow{3}{*}{Phan et al., 2022~\cite{sleeptrans}}                   & \multirow{3}{*}{EEG +~Transformer}                                               & \multirow{3}{*}{194}                                      & \textit{Pre}  & -             & -             & -             & -             & -             & \multirow{3}{*}{0.79}  & \multirow{3}{*}{0.85}\\
                            &                                                      &                                                                                           &                                                           & \textit{Re}   & -             & -             & -             & -             & -             &                                 \\
                            &                                                      &                                                                                           &                                                           & \textit{F1}   & \textbf{0.94}  & 0.49          & 0.87          & 0.81          & \textbf{0.85} &                                 \\ 
\cline{2-12}
                            & \multirow{3}{*}{Fiorillo et al., 2021~\cite{Fiorillo}}              & \multirow{3}{*}{EEG + CNN}                                                                & \multirow{3}{*}{98}                                       & \textit{Pre}  & 0.90          & 0.48          & 0.80          & 0.87 & 0.79          & \multirow{3}{*}{0.80}        & \multirow{3}{*}{0.85}   \\
                            &                                                      &                                                                                           &                                                           & \textit{Re}   & 0.93 & 0.44 & 0.85          & 0.73          & 0.73          &                                 \\
                            &                                                      &                                                                                           &                                                           & \textit{F1}   & 0.91          & 0.46 & 0.83          & 0.79          & 0.76          &                                 \\ 
\cline{2-12}
                            & \multirow{3}{*}{Korkalainen et al., 2020~\cite{Korkalainen}}           & \multirow{3}{*}{EEG, EOG + CNN}                                                  & \multirow{3}{*}{153}                                      & \textit{Pre}  & 0.93 & 0.45          & 0.87          & 0.78          & 0.85 & \multirow{3}{*}{0.80}     & \multirow{3}{*}{\textbf{0.86}}      \\
                            &                                                      &                                                                                           &                                                           & \textit{Re}   & 0.90          & 0.32          & 0.86          & 0.76          & 0.83          &                                 \\
                            &                                                      &                                                                                           &                                                           & \textit{F1}   & 0.92          & 0.38          & 0.86          & 0.77          & 0.84          &                                 \\ 
\cline{2-12}
                            & \multirow{3}{*}{Qu et al., 2020~\cite{nn1}}                  & \multirow{3}{*}{EEG + CNN, Transformer}                                                  & \multirow{3}{*}{79}                                       & \textit{Pre}  & 0.85          & 0.49 & \textbf{0.89} & \textbf{0.89}          & 0.81          & \multirow{3}{*}{0.79}       & \multirow{3}{*}{0.84}    \\
                            &                                                      &                                                                                           &                                                           & \textit{Re}   & \textbf{0.96}          & 0.48          & 0.86          & 0.82 & \textbf{0.85} &                                 \\
                            &                                                      &                                                                                           &                                                           & \textit{F1}   & 0.90          & 0.48          & 0.88          & \textbf{0.86}          & 0.83          &                                 \\
\toprule
\end{tabular}
\end{table*}

\begin{figure}[t]
\centering
\includegraphics[width=0.93\linewidth]
{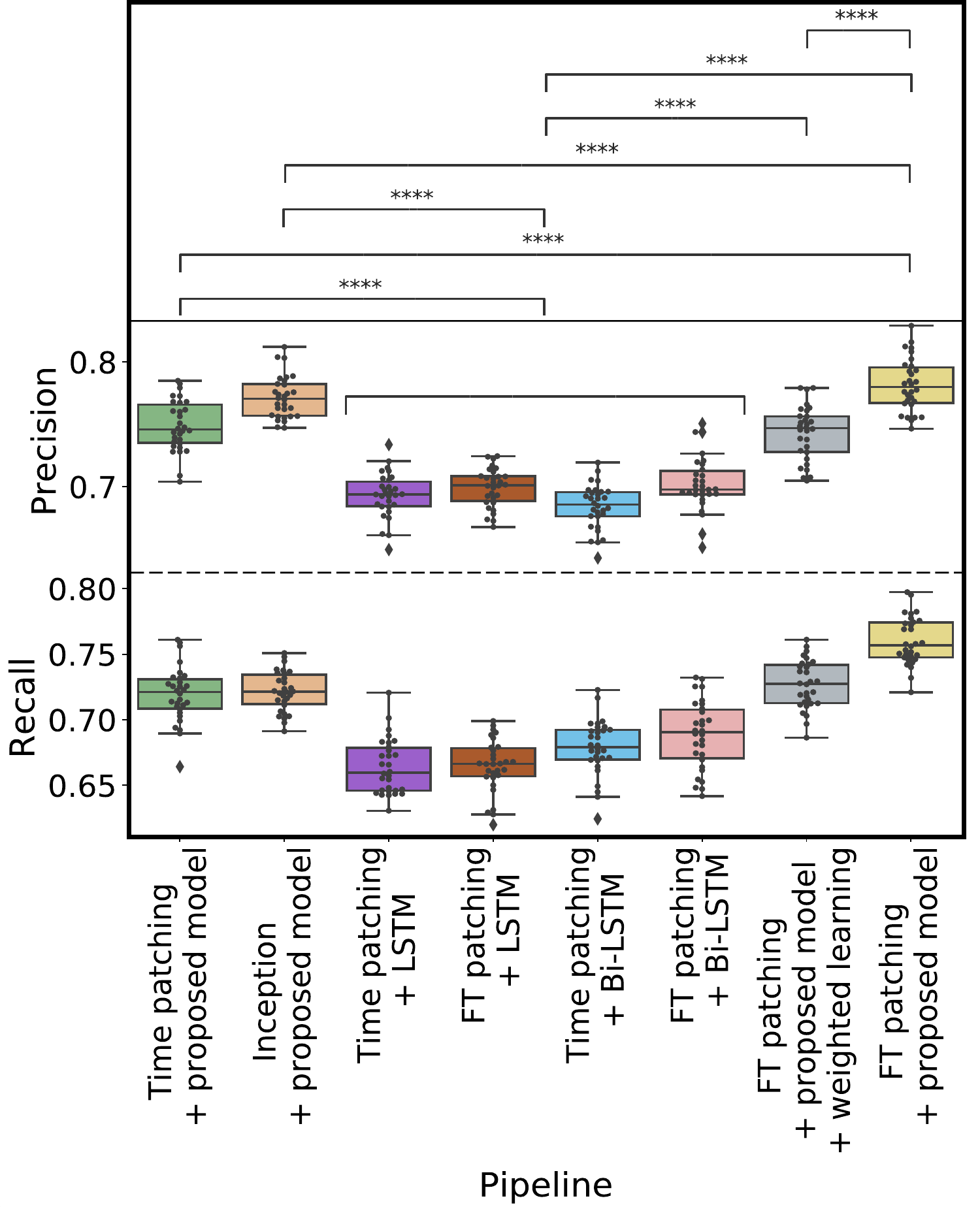}
\caption{Comparison of stage-wise performance among ablation methods and our proposal. 
The \textit{t}-test result illuminates both the Precision and Recall. The four asterisks **** mean $p < 0.0001$. For the \textit{t}-test we combine the RNNs-based models for the sake of a concise image.}
\label{fig:ablation}
\end{figure}

\textit{\textbf{Observation 1: The proposed model performs better in generating the stage-dependent features.}} Looking closer at the performance of the baseline models in Fig. \ref{fig:ablation}, two conclusions can be drawn. 
The first is straightforward as the models using the attention mechanism outperformed the RNN-based ones. The combination of FT patching and the Transformer model is significantly better than the others in terms of \emph{Pre} and \emph{Re}.
The second is that the best of RNN models comes from FT Patching+Bi-LSTM, which is different from the proposed framework in terms of the model architecture only, it is reasonable to conclude that the proposed model is a better architecture for generating stage-dependent features.

\begin{figure}[t]
\centering
\includegraphics[width=0.99\linewidth]{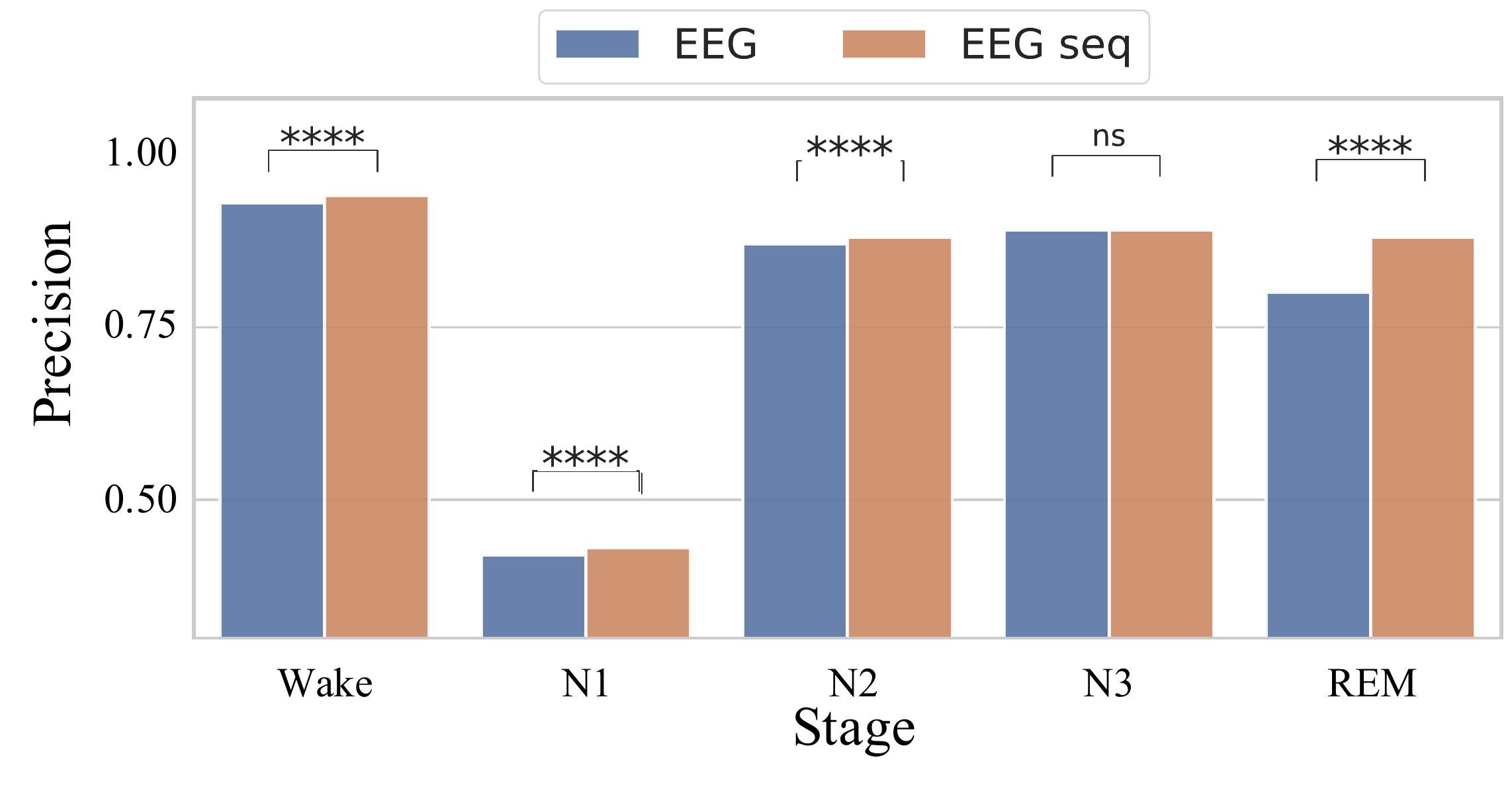}
\caption{Student's \emph{t}-test of sole EEG epoch to the EEG epoch sequence. 
**: $p < 0.01$; ***: $p < 0.001$.
}
\label{fig:t-test}
\end{figure}

\begin{figure*}[t]
\centering
\includegraphics[width=0.99\linewidth]{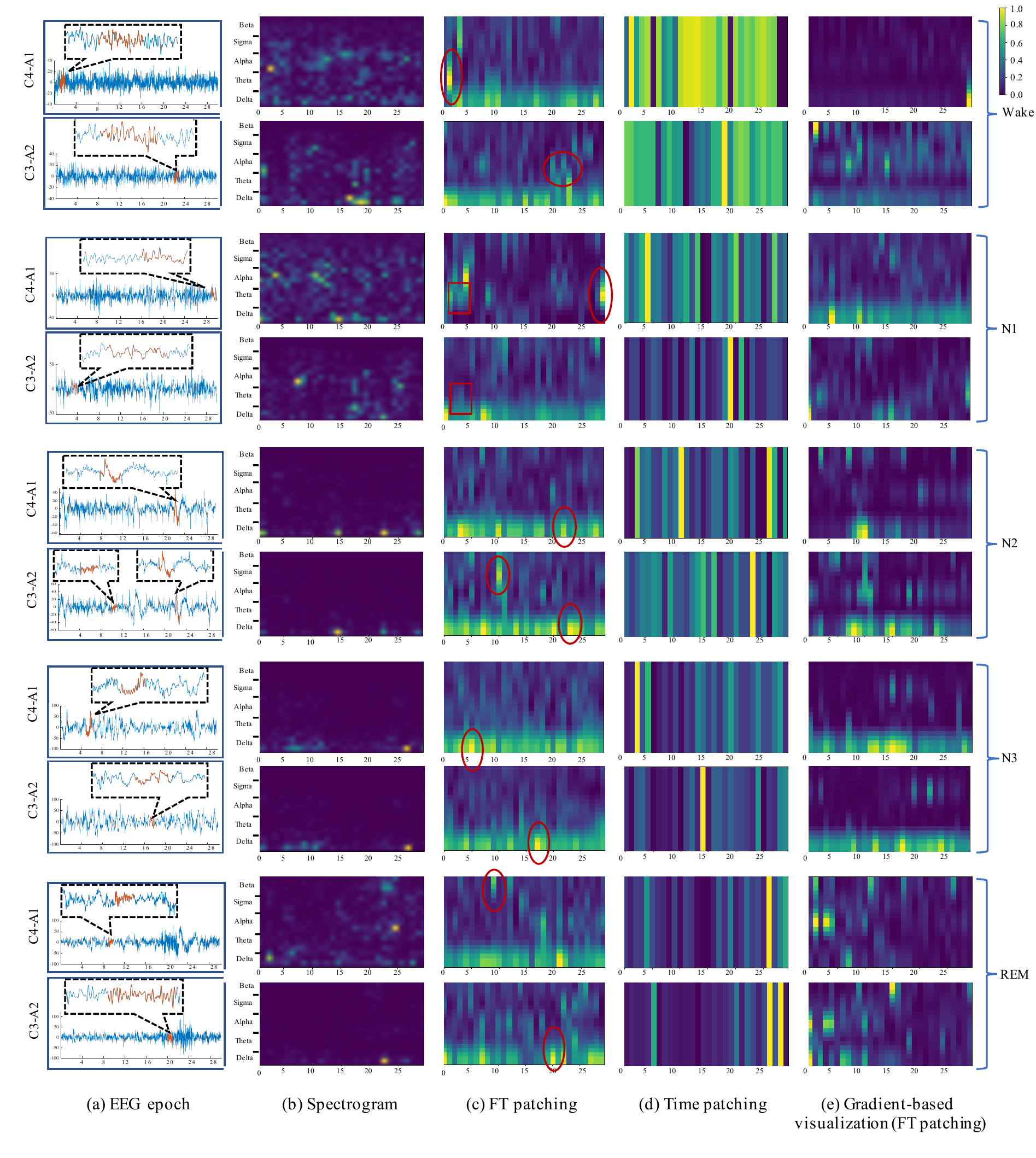}
\caption{Visualization of the attention-based proposal, time patching baseline, and gradient-based baseline.}
\label{fig:avv}
\end{figure*}

\textit{\textbf{Observation 2: The proposed frequency-time patching is an appropriate representation of sleep stages.}} 
The conclusion is drawn by comparing the model performance with different inputs, namely the time patching and FT patching of the spectrogram and the raw EEG signal input to the Inception module. Among the three ablations, retaining the frequency-band information of the Inception/Patching + proposed model showed its effectiveness compared to the time patching+proposed method in terms of overall performance. 
Given that the design of the \emph{Inception} module served the same purpose of retaining the resolution in the frequency domain, we believe the combination of spectrogram and FT patching is an appropriate representation of sleep stage-relevant information.

\begin{figure*}[t]
\centering
\includegraphics[width=0.99\linewidth]{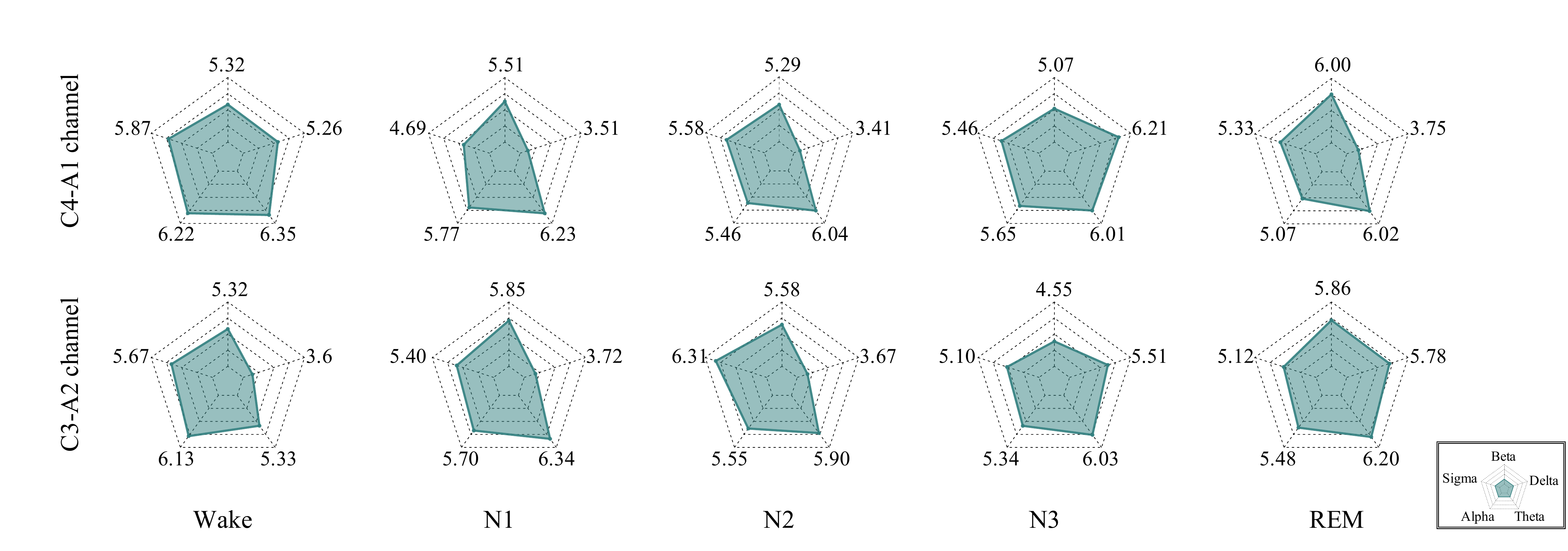}
\caption{Visualization of the attention-derived entropy of each frequency band of the proposed framework.
}
\label{fig:avv_entro}
\end{figure*}

\begin{figure*}[t]
\centering
\includegraphics[width=0.99\linewidth]{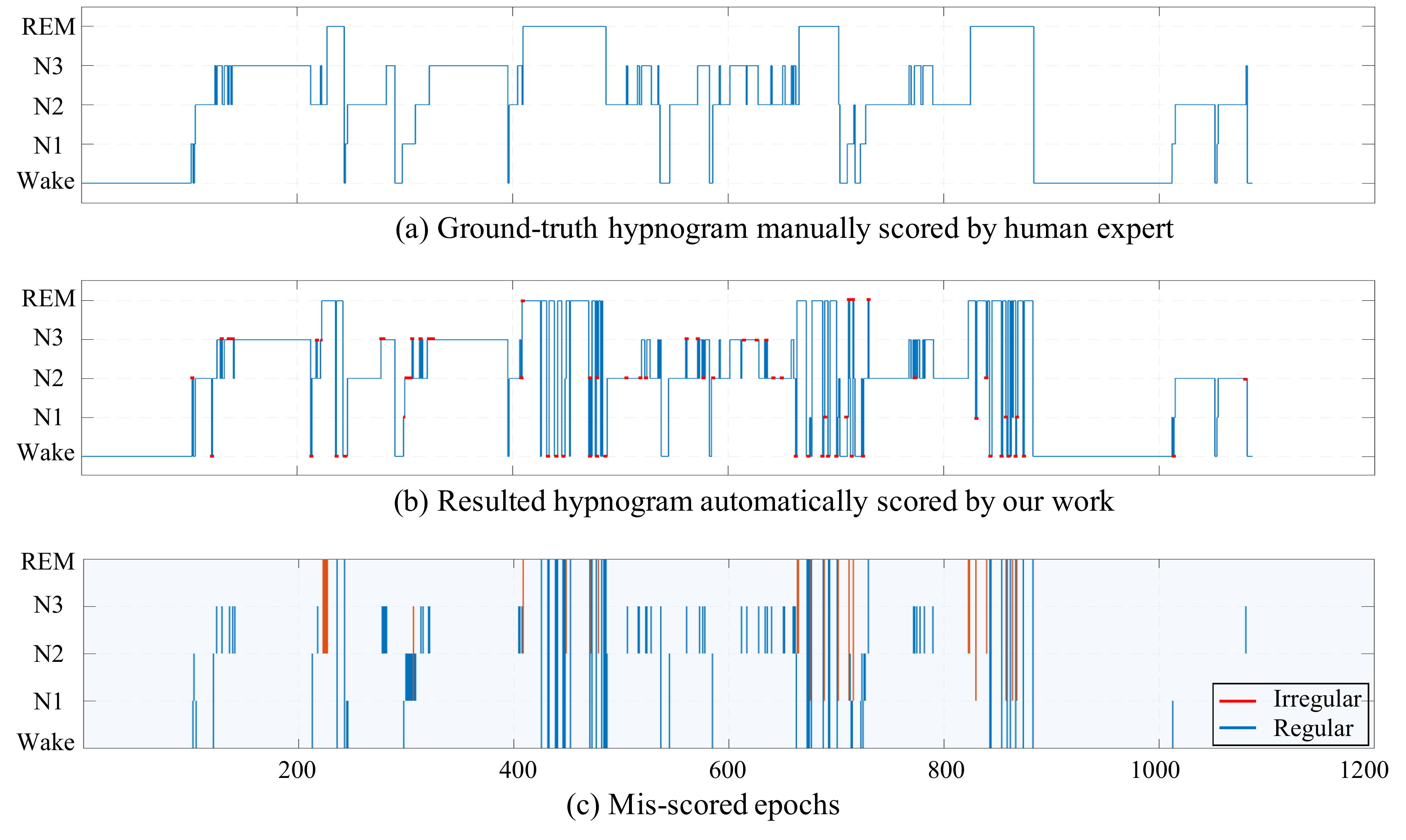}
\caption{Examples of hypnogram manually scored by the human expert (a) and hypnogram automatically scored by our method (b) for one subject from the SHHS dataset. Misclassification is marked in red. The sticks in the bottom figure (c) mark the wrong labels. The blue sticks represent the regular sleep stage transitions that can not be detected; while the red sticks represent the falsely detected irregular transitions.}
\label{fig:hyponogram}
\end{figure*}

\textit{\textbf{Observation 3: The proposed method achieved a new SOTA for almost all stages.}} 
We compared the proposed method against related works in TABLE \ref{sota}. We can observe that the classification of our method on the wake stage had the best performance in the SHHS dataset. 
The proposed method achieved 0.85 and 0.89 overall \emph{Acc} in single epoch and sequence staging strategy, respectively.
Meanwhile, the proposed method also outperformed other methods on stages N2 and N3, with 0.90 of \emph{Re} and 0.89 of \emph{Pre}, respectively.
Similar observations also resulted in the Sleep-EDF dataset, we hence achieved the optimal \emph{Acc} within 0.88 against the previous works.
The sequence Transformer proposed in \cite{sleeptrans} applies the Transformer to inter-epoch EEG signals and attains the highest performance on other metrics for N2 and N3.
By extending the framework into the EEG sequence (described in Section \ref{sec:related}), the performance can be further lifted for every sleep stage (refer to Table \ref{sota} and Figure \ref{fig:t-test}).
Pathak et al. \cite{ShreyasPATHAK} reached the highest performance for the REM stage by fusing the EOG and EMG signals, where those two kinds of signals are commonly considered important for the REM stage. 
Compared to the multi-resource works in \cite{shuiwang} and \cite{nn1}, the proposal has high inter-rater reliability with 0.85 and 0.81 \emph{k}-score in the SHHS and Sleep EDF database, respectively.

As will be mentioned in the Discussion section, fusing EOG signals into our proposal also obtained the new SOTA performance on stage REM, but in this paper,
our focus is on EEG. 
We also demonstrated competitive \emph{Pre} score by leveraging the sequence-to-sequence training strategy. 
Similar results can be found in the evaluation on top of the Sleep-EDF dataset.
Note that our proposal attained the best performance for N1 (\emph{Re}: 0.51 and \emph{F1}: 0.50) without using another signal resource, such as EOG.
Focus on our proposal, the single-epoch-based method also had a competitive overall performance against related works.

\textit{\textbf{Observation 4: FT patching together with the attention mechanism can better capture sleep-related features in parallel.}} To discuss the interpretability of the proposed method, we visualized the attention scores of FT patches/time patches input in Fig. \ref{fig:avv}. 
For simplicity, we normalized the intensity of each reconstructed spectrogram-like attention map. 
Gradient-based visualization (GbV) using Grad-CAM [column (e) in Fig. \ref{fig:avv}] was also generated for a direct comparison.

\textbf{Band Contribution. }
Regarding the wake stage, the bright patch around the 2nd second of the spectrogram (C4-A1 channel) of the $\alpha$ band has the correspondence in the FT patch map (see the stride in the ellipse), and the bright patches of the $\alpha$ band can also be found around 22nd second in C3-A2. 
By contrast, solely focusing on the time domain as shown in Fig. \ref{fig:avv} (d) can fail to extract critical information for determining the stages.
For the N2 stage, most of the K-complex shown in both EEG signals and spectrogram was closely attended to. 
Besides, a spindle-like patch (see the red box in the C3-A2 channel) was given high attention as well. 

\textbf{Attention vs GbV. }
The existing method GbV suggests different FT patches [bright spots in Fig. \ref{fig:avv} (e)] contributed to the identification of the sleep stages. 
However, some of the highlighted FT patches were not supported by clinical findings such as the bright strides in the $\delta$ band of C4-A1 for stage wake and $\delta$ band of stage N1.
By contrast, the attention map of FT patches resulted in a distinct view of the input that is better supported by the prevailing sleep knowledge.

\begin{figure}[t]
\centering
\includegraphics[width=0.98\linewidth]{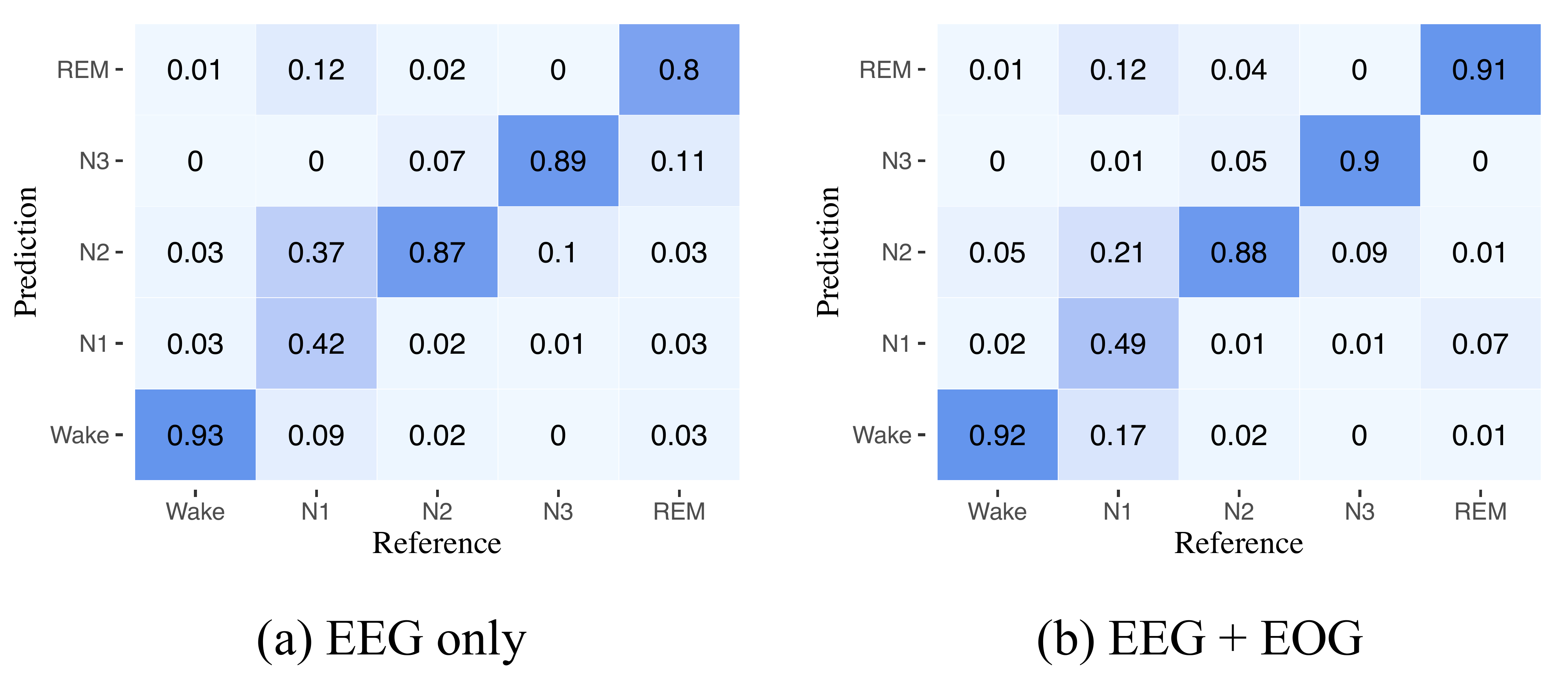}
\caption{Confusion matrix of the proposed method with EEG as the input (a) and the proposed method with EEG and EOG as the input (b).
}
\label{fig:eeogg}
\end{figure}

\begin{figure}[t]
\centering
\includegraphics[width=0.98\linewidth]{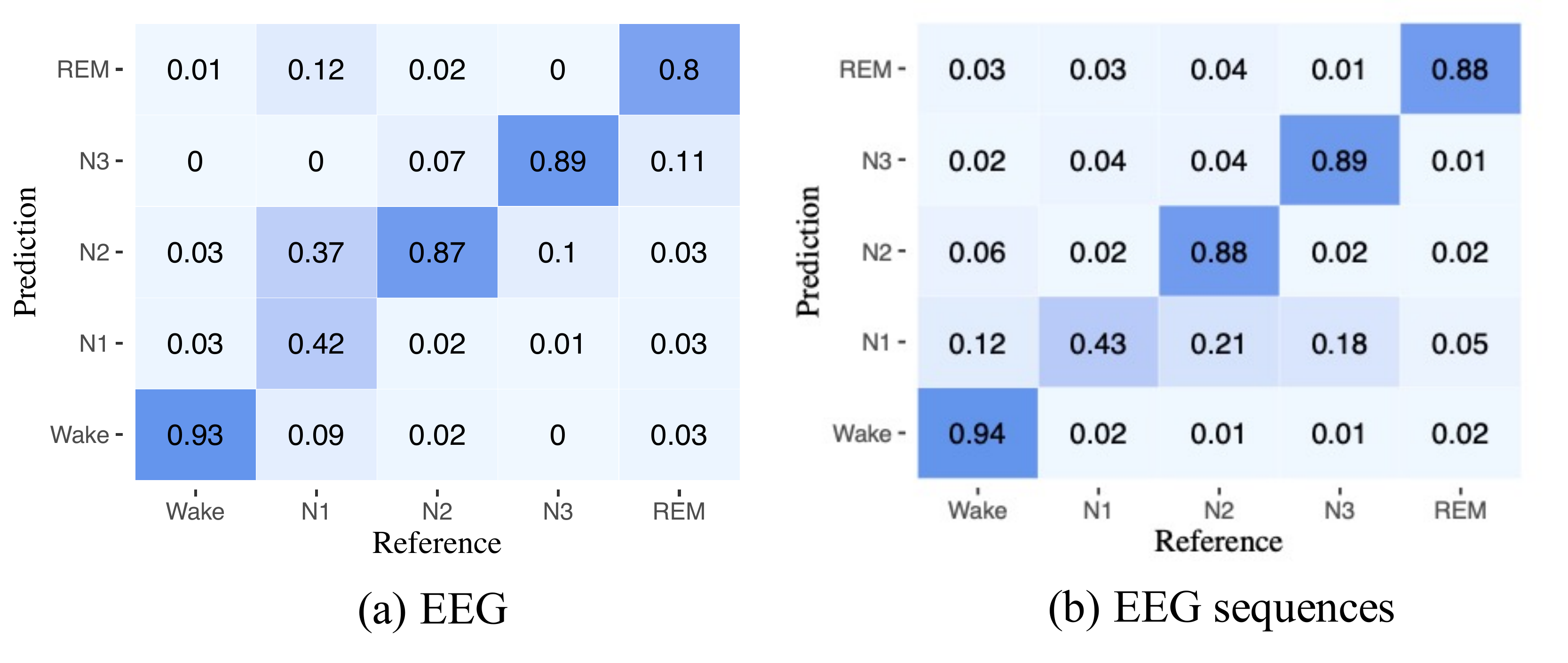}
\caption{Confusion matrix of EEG epoch (a) and EEG sequence (b).
}
\label{fig:eeggs}
\end{figure}

\section{Discussion}\label{sec:discussion}

Fundamentally, this paper aims to push forward epoch-wise automatic sleep staging using EEG signals to new highs with accurate sleep staging.  
In the meantime, the pipeline in this method should be an interpretable framework to capture the representative stage-specific features that are in accordance with the EEG characteristics defined by the sleep community while making an accurate staging decision.
From the experimental results, it is reasonable to conclude that these purposes have been fulfilled. 

We additionally included an entropy-based statistical analysis in the radar graphs Fig. \ref{fig:avv_entro}. 
They quantify the causality between the attention visualization and the model decision.
Considering the transient nature of stage-dependent features, the attention intensities in one frequency band distributed homogeneously might contain useful information.
Otherwise, lower intensity should have led to a lower sample entropy value. 
In the stage wake, the entropy of the $\alpha$ band reached relatively higher values with 6.21 and 5.51 in the C4-A1 and C3-A2 channels.
From wake to N1, the dominant $\alpha$ band attenuated in N1, accompanying the increase in the $\theta$ band. 
As the sleep went deeper, the attention given to the $\alpha$ band became stable at a relatively low pace, and the $\delta$ band gradually came into the foreground. 

By comparing clinical annotations of sleep stages against the automated staging of our proposal (Fig. \ref{fig:hyponogram}), it is visible that the proposal yielded concord with the clinical annotation in general.
However, there are some misclassifications that occurred between the \{N2, N3\} pair and the \{Wake, REM\} pair shown in TABLE \ref{finalpairs}. 
That is, the proposed method was upset when the sleep stages were transitional frequently in relatively short intervals (see the red dots in the middle hypnogram of Fig. \ref{fig:hyponogram}).
The reason might be the incompleteness of sleep-relevant information in the EEG signal \cite{N1hard}, the entangled stage-irrelevant information from the neighbor stage hence taking in dominant.
Although our model can recognize transitions of stages with relatively low frequency accurately, improving the sensitivity of stage transitions is our future work.
Furthermore, irregular misclassification pairs that the inter-epoch transitions violated the regular sleep cyclic pattern can be seen, occupying about 18\% of the total misclassification.
For instance, our proposal may output assignments of \{N1, REM\} ($N1\rightarrow REM$ or $REM\rightarrow N1$), a sharp change of stage that skips the intermediate N2 and N3 stages.
For the irregular pairs of \{N2, REM\}, N2 sometimes changed to REM without the deep sleep phase (around 200 epochs in Fig. \ref{fig:hyponogram}). 
Such change seldom happened once the body was stable at the REM stage. 
This issue suggests introducing constraints on inter-epoch relationships is necessary for future work.

We now further discuss the additional experiments to improve the staging performance of stage REM in two manners.
Since the EOG is considered indispensable for identifying the stage REM, we showed the experimental results on EEG merged with EOG in Fig. \ref{fig:eeogg}.
Although there is a significant improvement, any additional sensor involved inevitably leads to the influence of natural sleep in practice. 
Therefore, in this work, we restricted our proposal to EEG signals and analyzed the feasibility of extended at-home uses, where the simplicity of the sensor attachment has priority. 
Phan., et al proposed several sequence-to-sequence frameworks that receive a sequence of multiple epochs as input and make the staging by merging sequential information \cite{shuiwang, sp4, sleeptrans}.
Such methods exhibited the potential for staging REM by sole EEG, and our extensive experiment (Fig. \ref{fig:eeggs}) showed the accuracy gap between EEG and EEG-seq was most pronounced at REM with P values < 0.0001 from the t-test shown in Fig. \ref{fig:t-test}.
However, this framework leads to ambiguity in the modeling of transient stage-dependent features within intra-epochs; meanwhile, the sequence-to-sequence classification strategy inevitably creates a substantial memory overhead in implementation \cite{CHENmethods}.
In summary, another interesting future direction includes  investigating how the performance of staging REM could be further improved by more advanced methods.

\begin{table}[t]
\centering
\caption{Different types of misclassifications and their counting results. Each pair illuminates the two-direction inter-epoch transitions, e.g., \{Wake, N1\}: $Wake\rightarrow N1$ and $N1\rightarrow Wake$.}
\begin{tabular}{|lc|cc|}
\hline
Regular pairs & \#Pair & \multicolumn{1}{l}{Irregular pairs} & \#Pair \\ \hline
\{Wake, N1\}            & 12     & \multicolumn{1}{l}{\{N1, REM\}}               & 9      \\
\{Wake, N2\}            & 6      & \multicolumn{1}{l}{\{N2, REM\}}               & 18     \\
\{N1, N2\}              & 14     &\multicolumn{1}{l}{\{Wake, N3\}}                                          & 1      \\
\{N2, N3\}              & 58     & -                                             & -      \\
\{Wake, REM\}           & 35     & -                                             & -      \\ \hline
\end{tabular}
\label{finalpairs}
\end{table}

\section{Conclusion}\label{sec:conclusion}
Developing automated sleep staging systems is an important effort toward facilitating the precise diagnosis of sleep-related diseases and serving neuroscientific explorations in the human brain. 
While previous studies have demonstrated the effectiveness of deep learning-based methods for this task, these approaches have not yet been fully optimized to align with the natural characteristics of sleep stage scoring.
This study introduced a novel feature processing framework that reflected the nature of the defining characteristics of EEG signals and proposed a novel attention-based sleep staging model. 
The effectiveness of our feature extraction process and model architecture was validated using a large-scale database.
Our model evaluation showed that the proposed method achieved the best performance for the wake, N2, and N3 sleep stages. 
Our findings also suggest that attention-based models may be particularly well-suited for representing the transient characteristics of sleep stages in parallel. Overall, this study represents an important step forward in the development of automated sleep staging systems and has significant implications for sleep medicine and neuroscience research.

\bibliographystyle{IEEEtran}
\bibliography{IEEEabrv,reference}

\end{document}